\documentclass{article}

\PassOptionsToPackage{numbers}{natbib}

\usepackage{titletoc}
\usepackage{tocloft}
\usepackage[preprint]{neurips_2019}

\usepackage[utf8]{inputenc} 
\usepackage[T1]{fontenc}    
\usepackage{hyperref}       
\usepackage{url}            
\usepackage{booktabs}       
\usepackage{amsfonts}       
\usepackage{nicefrac}       
\usepackage{microtype}      

\usepackage{xspace}

\usepackage{amsmath,amssymb}

\usepackage{graphicx}
\usepackage{placeins}
\usepackage{tabularx}
\usepackage{adjustbox}
\usepackage{pifont}
\usepackage{multirow}
\usepackage{floatrow}
\usepackage{wrapfig}

\usepackage{color}
\usepackage{xcolor}

\newcommand{\vx}{\mathbf{x}}
\newcommand{\vxh}{\hat{\mathbf{x}}}
\newcommand{\vy}{\mathbf{y}}
\newcommand{\vz}{\mathbf{z}}
\newcommand{\vyh}{\hat{\mathbf{y}}}
\newcommand{\vzh}{\hat{\mathbf{z}}}
\newcommand{\vyhadv}{\hat{\mathbf{y}}_\textrm{adv}}
\newcommand{\vzhadv}{\hat{\mathbf{z}}_\textrm{adv}}

\newcommand{\vh}{\mathbf{h}}
\newcommand{\vhy}{\mathbf{h}_y}
\newcommand{\vhz}{\mathbf{h}_z}

\newcommand{\vW}{\mathbf{W}}
\newcommand{\vw}{\mathbf{w}}

\newcommand{\calL}{\mathcal{L}}

\DeclareMathOperator{\softmax}{SoftMax}
\DeclareMathOperator{\sigmoid}{sigmoid}

\definecolor{BrickRed}{HTML}{B6321C}
\definecolor{RoyalBlue}{HTML}{0071BC}
\definecolor{PineGreen}{HTML}{008B72}
\definecolor{bluefig}{HTML}{5B9BD5}

\newcommand{\parag}[1]{\textbf{#1\hspace{1mm}}}

\newcolumntype{R}[2]{%
    >{\adjustbox{angle=#1,lap=\width-(#2)}\bgroup}%
    l%
    <{\egroup}%
}

\newcommand{\okI}{$\vy$}
\newcommand{\okA}{$\vz$}
\newcommand{\okAt}{$\vz_\textrm{eval}$}

\title{DualDis: Dual-Branch Disentangling with Adversarial Learning}

\usepackage{authblk}

\author[1]{Thomas Robert}
\author[2]{Nicolas Thome}
\author[1,3]{Matthieu Cord}
\affil[1]{Sorbonne Université, CNRS, Laboratoire d’informatique de Paris 6, LIP6, F-75005 Paris, France}
\affil[2]{CEDRIC, Conservatoire National des Arts et Métiers, 75003 Paris, France}
\affil[3]{valeo.ai, Paris, France}
\affil[ ]{\texttt{firstname.lastname@\{lip6,cnam\}.fr}}

\begin{document}

\maketitle

\begin{abstract}
In computer vision, disentangling techniques aim at improving latent representations of images by modeling factors of variation.
In this paper, we propose DualDis, a new auto-encoder-based framework that disentangles and linearizes class and attribute information. This is achieved thanks to a two-branch architecture forcing the separation of the two kinds of information, accompanied by a decoder for image reconstruction and generation. To effectively separate the information, we propose to use a combination of regular and adversarial classifiers to guide the two branches in specializing for class and attribute information respectively. We also investigate the possibility of using semi-supervised learning for an effective disentangling even using few labels. We leverage the linearization property of the latent spaces for semantic image editing and generation of new images. We validate our approach on CelebA, Yale-B and NORB by measuring the efficiency of information separation via classification metrics, visual image manipulation and data augmentation.
\end{abstract}

\section{Introduction}

In deep learning and especially in the computer vision community, the problem of disentangling factors of variation~\citep{Mathieu2016,higgins2017beta} is a very active field of research. The overall objective of disentangling is to increase the quality of the latent representations so that they represent independent factors of variation in the data. This can be done in unsupervised~\citep{chen2018isolating,Hu2018,dupont2018learning}, weakly supervised~\citep{szabo2017challenges,Feng2018,ruiz2019learning} or supervised ways~\citep{Wang2017,Jaiswal2018} and can have a wide range of applications. By improving the quality of the representations, these can be used to improve transfer learning~\citep{ruiz2019learning}, domain adaptation~\citep{chang2019all,louizos2016}, information retrieval~\citep{Mathieu2016}, \textit{etc.} In addition, since these models are usually based on encoder-decoder architectures~\citep{bourlard1988auto,hinton2006reducing}, they can combine \textit{visual understanding} and \textit{image generation}. While disentangling models do not directly compete with generative models~\citep{Goodfellow2014,zhu2017unpaired,Bodla2018,karras2017progressive,wang2018high,choi2018stargan,karras2018style}, they can be an interesting direction for controlling latent conditional factors regarding what is being generated, which remains a challenging task in this literature.

In this paper, we want to explore structured latent representations of images designed for \textit{image classification} and \textit{visual attribute detection}. We are interested in modeling two complementary kinds of information that we will call \textit{information domains}. For example, with a face dataset, we would like to represent the identity (\textit{i.e.} the class) of the person and various visual attributes (hairstyle, makeup, facial expression, \textit{etc.}).

\begin{figure}[tb]
    \centering
    \includegraphics[width=\linewidth]{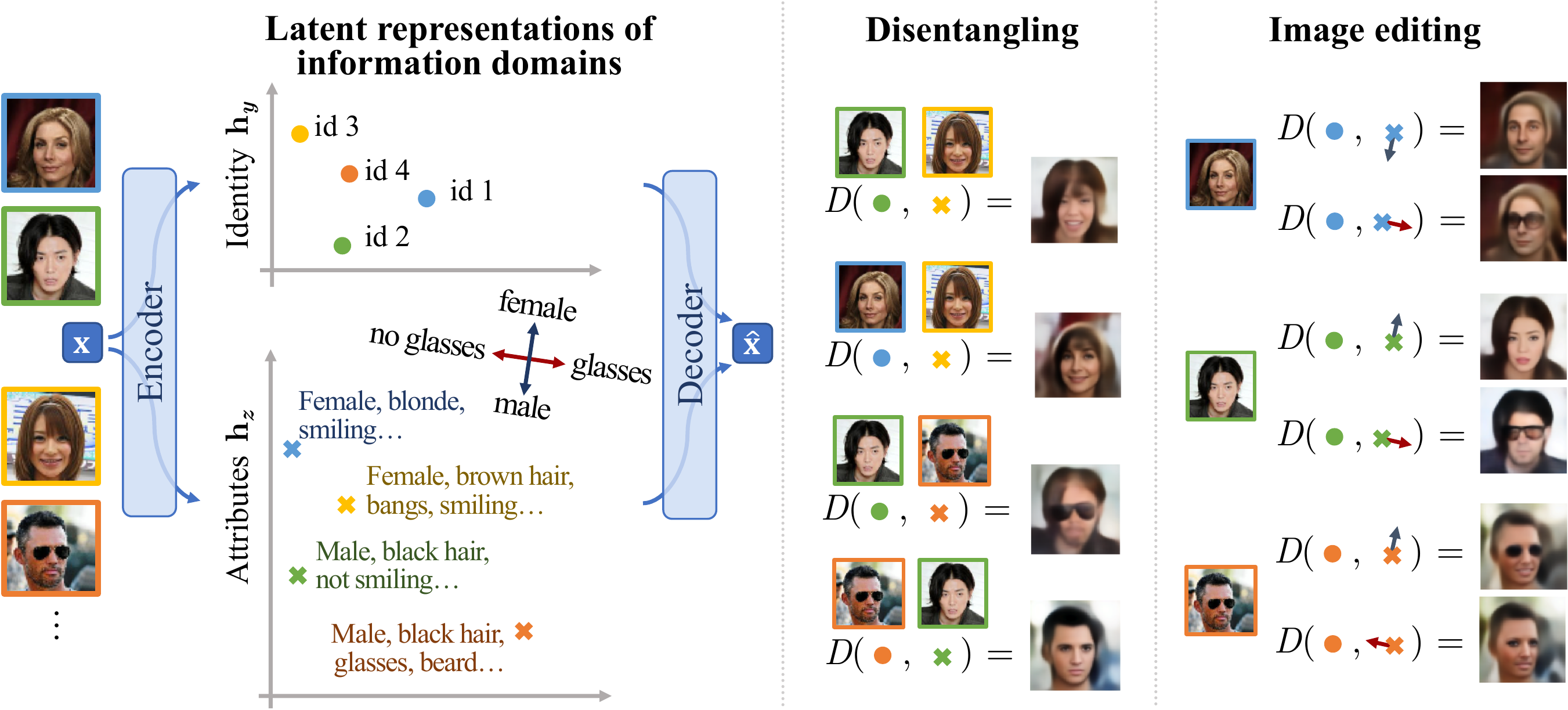}
    \caption{Overview of our DualDis framework.
    On the left we illustrate the behavior of our encoder-decoder, learned to explicitly separate complementary representations of identity (top) and attributes (bottom) in dual latent subspaces.
    In the middle, we illustrate its \textit{disentangling} ability by being able to mix the identity of a first image and the attributes of a second. In the first example, the green man takes the attributes of the yellow image, becoming a smiling woman with brown bangs.
    As our model also linearizes the factors of variation, one can perform \textit{image editing} (right). For the first example (blue woman), we move the representation \textcolor{bluefig}{\scriptsize \ding{54}} along the directions male (first line) and glasses (second line) to add those attributes.}
    \label{fig:motivation}
\end{figure}

To achieve this goal, we introduce a deep architecture that will explicitly separate the information domains in two distinct latent subspaces (Fig.~\ref{fig:motivation}, left). The first latent space $\vhy$ encodes only class-related information while the second $\vhz$ encodes attribute-related information; a specialization guided by classification losses. A decoder $D(\vhy, \vhz)$ is used to reconstruct images and generate new ones.
Our  main contribution is the learning strategy that we propose to train this architecture.
Using adversarial training, we are able to explicitly separate and ``orthogonalizes'' the information from the two information domains. To achieve this, each latent space is connected to a classifier of the opposite information domain, so that this classifier finds the information that belongs to the wrong domain. The encoder will then learn to remove this information from the latent space, making classification impossible and thus filtering only the relevant information. Our approach is called DualDis to highlight our two branch disentangling process. It is illustrated in Fig.~\ref{fig:motivation} (middle, ``disentangling'') where it is possible to mix representations of different images.
We study the disentangling capabilities of our model on CelebA~\citep{celeba}, Yale-B~\citep{yale} and NORB~\citep{norb} by comparing our model quantitatively to state-of-the-art models by measuring both the accuracy of the models for identity and attributes classification and their ability to disentangle the two information domains.

In addition, our architecture is also designed to linearize the factors of variation in each latent space, a behavior that reinforces the semantic quality of the representation and is necessary for effective image generation and editing. This is ensured through linear classifiers that both guide the linearization process and provide us with linear directions to semantically navigate the latent spaces. Thanks to this, we are able to modify the information represented by our latent variables that has been extracted from any given image and perform image editing.
For example on Fig.~\ref{fig:motivation} (right), we change the gender and eyeglasses attributes of images while conserving the identity and the other attributes. Thanks to this, we perform \textit{guided} data augmentation by generating variations of images with semantic changes instead of low-level changes (flip, translation, color jitter, \textit{etc.}) as usually done. We leverage this capability to significantly improve identity classification performance on the Yale-B face dataset.

\section{DualDis approach}

\begin{figure}[tb]
    \centering
    \includegraphics[width=\linewidth]{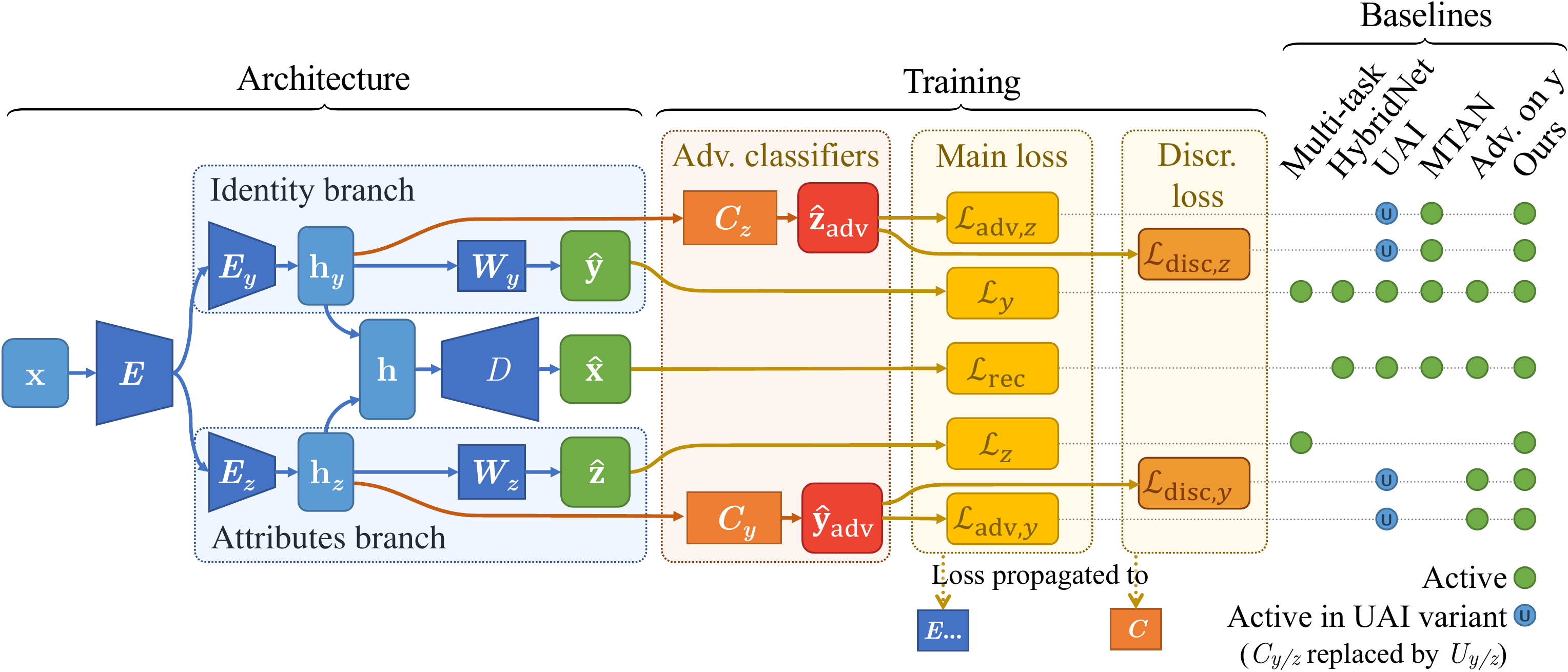}
    \caption{Architecture and learning of the DualDis framework. We use a two-branch encoder-decoder architecture with classifiers (left). For the training, in addition to classical reconstruction and classification losses, we use adversarial classifiers and loss terms to force the two latent spaces to encode complementary ``orthogonal'' information (middle).
    We also indicate (right) how subsets of components of DualDis can be used to reproduce existing baselines described in Sec.~\ref{sec:discussion}.}
    \label{fig:archi}
\end{figure}

We propose an approach called DualDis presented on Fig.~\ref{fig:archi}. On the left, we show the architectural part of our contribution, using disentangling to separate two \textit{information domains} (class/identity and attributes). Those domains have classification labels $\vy$ and $\vz$ that we want to predict ($\vyh$, $\vzh$), along with a reconstruction $\vxh$ of the input $\vx$. In the center of the figure, we describe the second part of our approach which is the training process designed to successfully disentangle the two domains, using adversarial classifiers and multiple loss terms. On the right, to put our model in perspective, we indicate how some related works can be reproduced using the same kind of architectures with variations in the losses used.

\subsection{Dual branch Auto-Encoder}\label{sec:model_ae}

We propose an encoder-decoder architecture with a latent space split in two parts, $\vhy$ and $\vhz$. Each representation is produced by a deep encoder $E_y$ or $E_z$ so that the features are explicitly separated. These representations are concatenated into $\vh$ and fed to a decoder $D$, producing a reconstruction $\vxh$. Having a decoder enables image generation and ensures that the model extracts robust features~\citep{le2018supervised}.

While it would be possible to encode all the information in a single latent space, having two branches encourages the model to encode two complementary kinds of information~\cite{Mathieu2016,Hadad2018,Liu2018}. Taking the example of a face dataset, we want the identity branch ($E_y \circ E$) to capture information related to the identity $\vy$ with invariance toward other factors of variation (hair style, makeup, pose, \textit{etc.}); and we want the attribute branch ($E_z \circ E$) to model this ignored information, since this branch needs to capture factors of variation linked to visual attributes $\vz$. Having two separate deep encoders $E_y$ and $E_z$ is key to an effective disentangling, and they should be designed deep enough to produce ``orthogonal'' latent representations that encode very different information. Since the low-level features represented by the first convolutional layers are likely common to both domains, we use a single common encoder $E$ before specializing the information in our two branches.

This auto-encoding backbone is trained using a simple mean-squared error, $\calL_\textrm{rec} = ||\vx - \vxh||_2^2$. A visual GAN discriminator~\citep{Goodfellow2014} or a perceptual loss~\citep{dosovitskiy2016generating} could improve the quality of the generations but this was not used since it is out of the scope of our paper.

\subsection{Modeling factors of variation}\label{sec:model_factors}

We want our architecture to produce robust representations of each information domain as well as provide classification predictions. First, we can note that having a two-branch encoder was shown to improve classification performance~\citep{Robert2018} by encouraging representations $\vhy$ and $\vhz$ to be invariant toward intra-class variations. To the encoder, we add linear classifiers $\vW_y$ and $\vW_z$, one for each branch, that predict respectively $\vyh$ and $\vzh$. These classifiers guide the auto-encoding backbone to organize the information extracted for reconstruction in the right branch between our two latent spaces $\vhy$ and $\vhz$ so that it allows to predict the class/identity and the attributes. To train those classifiers, we use regular classification losses. We have:
\begin{align}
    \vyh &= \softmax(\vW_y \vhy) \,, & \vzh &= \sigmoid(\vW_z \vhz) \,, \\
    \calL_{y} &= \textrm{CrossEntropy}(\vy, \vyh)         \,, &
    \calL_{z} &= \textrm{BinaryCrossEntropy}(\vz, \vzh)   \,.
\end{align}

\parag{Factors linearization and manipulation.} In addition, those classifiers are designed to encourage the linearization of the representations of labeled factors of variation. Indeed, the presence of the $i^\text{th}$ attribute $\vz_i$ in an input image is estimated by a linear predictor: $\vzh_i = \sigmoid(\vw_{z_i} \cdot \vhz)$ with $\vw_{z_i}$ the $i^\text{th}$ row of the matrix $\vW_z$.
This means that we can manipulate the latent space to artificially increase or decrease the presence of this attribute by moving $\vhz$ in the direction of this vector:
$\vhz' = \vhz \pm \varepsilon \vw_{z_i}^\top$.

\subsection{Disentangling information domains and factors of variation}\label{sec:model_dis}

We also want our architecture to explicitly disentangle the two domains. To this end, we add classifiers $C_y(\vhz)$ and $C_y(\vhy)$ that predict the target of the opposite domain ($\vy$ from $\vhz$ and vice versa). We call those classifiers ``adversarial'' since their role is to find information that should not be present, \textit{e.g.} information about attributes in the identity branch. Training a classifier to find this information allows the encoder to remove it. Those classifiers are designed as non-linear multi-layer classifiers to be able to find the information even if it is not linearly separable in the latent space.

\parag{Global loss and adversarial training.} We use adversarial training and thus have two losses that are backpropagated to different parts of the model, \textit{cf.} Fig.~\ref{fig:archi}. The \textit{main loss} $\calL$ describes the expected behavior of the model and is propagated in the encoders, decoder and classifiers $\vW$. The \textit{discriminative loss} $\calL_\textrm{disc}$ is used to train the adversarial classifiers $C_y$ and $C_z$ only; helping the main loss by searching for information that should be removed.
\begin{align}
    \calL\,(\theta_{E,E_y,E_z,D,\vW_y,\vW_z}) &= \lambda_r \calL_\textrm{rec} + \lambda_y \calL_{y} + \lambda_z \calL_{z} + \lambda_{a,y} \calL_{\textrm{adv},y} + \lambda_{a,z} \calL_{\textrm{adv},z} + \lambda_{o} \calL_\textrm{orth}\,,\\
    \calL_\mathrm{disc}\,(\theta_{C_y,C_z}) &= \lambda_{d,y} \calL_{\textrm{disc},y} + \lambda_{d,z} \calL_{\textrm{disc},z}\,.
\end{align}
We add weights $\lambda$ to control the importance of each loss term.

\parag{Disentangling information domains.}
To effectively remove the information that should not be encoded, we thus start with discriminative loss terms $\calL_{\textrm{disc},y}$ and $\calL_{\textrm{disc},z}$, applied to $C_y$ and $C_z$, to make the adversarial classifiers achieve correct classification of their target. We then use the terms $\calL_{\textrm{adv},y}$ and $\calL_{\textrm{adv},z}$ in the main loss to make the encoders produce features that prevent the classifiers $C$ to achieve their goal and ideally have the accuracy of a random classifier.
This goal can be expressed in multiple ways:
by maximizing the entropy of the prediction $H[\vzhadv]$;
by minimizing the cross-entropy of the prior distribution, \textit{e.g.} uniform among classes $\textrm{CrossEntropy}(\mathcal{U}_{N_y}, \vyhadv)$;
or by encouraging the prediction of the inverse of the ground truth, which we found to be the most effective:
\begin{align}
    \calL_{\textrm{disc},y} &= \textrm{CrossEntropy}(\vy, \vyhadv) \,,    &
    \calL_{\textrm{disc},z} &= \textrm{BinaryCrossEntropy}(\vz, \vzhadv) \,;\\
    \calL_{\textrm{adv},y} &= -\textrm{CrossEntropy}(\vy, \vyhadv) \,,   &
    \calL_{\textrm{adv},z} &= \textrm{BinaryCrossEntropy}(1-\vz, \vzhadv) \,.
\end{align}

\parag{Intra-branch disentangling.}
Using adversarial classifiers, we disentangled class and attribute features. We now propose to disentangle the different labeled factors of variation by making the rows of the matrix $\vW_z$ orthogonal, meaning each factor detector is independent from the others. We do so by minimizing the dot products of the pairs of normalized row vectors $\vw_{z_i}'$ of $\vW_z$:
\begin{equation}
    \calL_\mathrm{orth} = \sum_i \sum_j \vw_{z_i}' \vw_{z_j}'^{\top} \quad\text{ with }\quad \vw_{z_i}' = \frac{\vw_{z_i}}{||\vw_{z_i}||_2} \,.
\end{equation}

\parag{Semi-supervised learning.}\label{sec:model_ssl}
Finally, while our model relies on more labels than often used for disentangling, it is possible to minimize the annotation requirement by using semi-supervised learning (SSL). Ideally, only using few labeled samples would be sufficient to guide the disentangling process. A simple semi-supervised strategy is to ignore unlabeled samples whenever the label is necessary in a loss term, which only limits classification-based loss terms. To go further, we can use existing SSL techniques like the ones producing ``virtual'' targets~\citep{tarvainen2017mean,Laine2016,miyato2018virtual} or label propagation~\citep{iscen2019label} to improve classification performances and obtain estimated labels to use as targets whenever needed.

\newcommand{\MTref}{(A)\xspace}
\newcommand{\MTANref}{(C)\xspace}
\newcommand{\HNref}{(B)\xspace}
\newcommand{\HNpref}{(B')\xspace}
\newcommand{\HNrefs}{(B\,\&\,B')\xspace}
\newcommand{\UAIref}{(D)\xspace}
\newcommand{\UAIrefs}{(D\,\&\,D')\xspace}
\newcommand{\UAIpref}{(D')\xspace}
\newcommand{\Yref}{(E)\xspace}

\section{Related work}
\label{sec:discussion}

We discuss our DualDis approach in regard to related works and existing state-of-the-art models.
Interestingly, as shown in Fig.~\ref{fig:archi} (right), several existing disentangling models can be reproduced and evaluated using the same architecture and loss components. By disabling parts of our architecture and/or with small additions and variations, we can thus produce a complete and fair comparison to them. We assign letters (A) to (E) to those models as we will refer to them in the experiments.

\parag{Class-only adversarial disentangling.} First, by starting with DualDis and removing losses related to the attribute labels $\vz$, we obtain \textit{model~\Yref} that learns to predict the class $\vy$ in one branch and dispel the information related to $\vy$ in the other~\citep{Hadad2018,Liu2018a,Klys2018}. The advantage is that fewer annotations are required, however, this means that the disentangling is asymmetrical since attribute-related information can be encoded freely in $\vhy$ and $\vhz$ and can thus remain entangled. It is therefore very difficult to ensure a proper separation of the two information domains with this asymmetry.
In addition, without labels $\vz$, we lack a way to semantically navigate in the latent space $\vhz$. This is why we choose a strong supervision with labels for both $\vy$ and $\vz$.

\parag{Symmetric latent adversarial disentangling.} To solve this asymmetry, UAI~\citep{Jaiswal2018} proposes a new adversarial training used as \textit{model~\UAIref} still without attribute labels. Adversarial predictors $U_y$ and $U_z$ replace $C$ and learn to predict $\hat{\vh}_y$ from $\vhz$ and vice versa, \textit{i.e.} the opposite latent space. The encoders try to fool those predictors. It has the advantage of being symmetrical even without attribute labels, but only ``orthogonalize'' $\vhy$ and $\vhz$ without guarantee of semantic disentangling. In particular, nothing ensures attribute information is removed from $\vhy$ and identity information can even remain in $\vhz$ if absent from $\vhy$. This is why we choose an explicit disentangling of labeled information.

\parag{Attribute-conditional decoder.} Some models~\citep{Lample2017,perarnau2016invertible}, like MTAN~\citep{Liu2018} \textit{(model~\MTANref)}, propose to use a single latent space $\vhy$ to encode the identity and feed the decoder with a binary vector $\vz$: $D(\vhy, \vz)$. MTAN applies a classification loss on $\vy$ and an adversarial loss on $\vz$ to remove the attribute information from $\vhy$.
However, this makes the strong assumption that all the attribute information of a specific image is encoded in the binary vector $\vz$, which is unlikely for complex semantic attributes, like a pair of glasses, a hat, facial expression, age, \textit{etc.} This is why we prefer to use two latent spaces.

\parag{Multi-Task Learning.} Finally, two-branch multi-task models could also lead to disentangling thanks to the specialization induced by classification, even without adversarial training. We can consider a classifier predicting both $\vyh$ and $\vzh$ like UberNet~\citep{kokkinos2017} \textit{(model \MTref)}; or a model like HybridNet~\citep{Robert2018} \textit{(model \HNref)}, that learns to classify $\vyh$ while encoding additional information in $\vhz$ used for reconstruction.
Compared to those methods, we choose to integrate an explicit disentangling process to effectively separate the information domains.

\parag{Unsupervised disentangling.} Other approaches address the problem of unsupervised disentangling~\cite{higgins2017beta,chen2018isolating,Hu2018,dupont2018learning}. The main difference with DualDis is that the usual motivation is to find latent units that each encodes an independent factor of variation, without labels. As such, it is not easy to represent a complex semantic factor (\textit{e.g.} facial expression, pose, hair style, \textit{etc.}) using a single unit. In addition, since no label is used, each latent unit must be visually interpreted by a human.

\parag{Generative models.}
Finally, our model can be analyzed in relation to generative models. While we do not claim to compete on image quality, the design of our model has interesting properties regarding generation.
First, the combination of our two-branch encoder and linearization process produces a latent space that has been organized, and allows semantic manipulation of existing images. This encoder structuring the information is important to ``bridge the gap'' between generative and discriminative tasks. It is still rare in generative models~\citep{Kingma2013,Dumoulin2016,Donahue2016} and is also a direction to help to prevent mode collapse~\citep{rosca2017variational,bang2018mggan}. Unlike approaches like~\citet{engel2018latent} that assume the generative model separated the identity and attribute information in the latent space without supervision, we choose to explicitly enforce a structure in the latent space. This provides a strong control on what is generated and avoid the rigidity of a conditional model that uses binary input attributes~\citep{perarnau2016invertible}.

\begin{table}[tb]
    \caption{Comparison to state-of-the-art models. We indicate the labels necessary in train ($\vy, \vz$) and to use the model ($\vz_\mathrm{eval}$). We measure the \textit{accuracy} of the classifiers $\vW_y(\vhy)$ and $\vW_z(\vhz)$ to predict the classes and attributes;
    and the \textit{disentangling quality} as the error rate (100 - accuracy) of the classifiers $C_y(\vhz)$ and $C_z(\vhy)$ indicating if information was correctly removed.
    Our \textit{aggregated metric} is an average of the four scores to indicate the overall performance. For all the scores, \textit{higher is better}.}
    \label{tab:sota}
    \centering
    \setlength{\tabcolsep}{6pt}
    \begin{tabular}{@{}ll@{\hspace{4pt}}ll@{\hspace{15pt}}c@{\hspace{20pt}}cc@{\hspace{13pt}}cc@{}}
        \toprule
    &&                             & Labels             & Aggr.  & \multicolumn{2}{c}{Accuracy} &  \multicolumn{2}{c}{Disentangling} \\
   && Model                        & used               & metric
                                & ${\scriptstyle \vh_{\scriptstyle y} \rightarrow} \vy$
                                & ${\scriptstyle \vh_{\scriptstyle z} \rightarrow} \vz$
                                & ${\scriptstyle \vh_{\scriptstyle z} \rightarrow} \vy_\textrm{adv}$
                                & $\!\!{\scriptstyle \vh_{\scriptstyle y} \rightarrow} \vz_\textrm{adv}$ \\
\midrule

\parbox[t]{3mm}{\multirow{8}{*}{\rotatebox[origin=c]{90}{\textbf{CelebA}}}}
& \MTref   & Multi-task classif.                & \okI, \okA                  &     61.1 & \bf 77.6\% & \bf 91.8\% &     65.5\% & \,\  9.5\% \\ 
& \HNref   & HybridNet-like~\citep{Robert2018}  & \okI                        &     65.1 &     73.0\% &     82.4\% &     95.5\% & \,\  9.4\% \\ 
& \HNpref  & HybridNet-like + attr              & \okI, \okA                  &     65.2 &     72.7\% &     90.1\% &     88.5\% & \,\  9.5\% \\ 
& \MTANref & MTAN~\citep{Liu2018}               & \okI, \okA, \okAt           &      --  &     68.9\% &        --  &        --  &     13.8\% \\ 
& \UAIref  & UAI  adv. loss~\citep{Jaiswal2018} & \okI                        &     63.7 &     67.9\% &     80.3\% & \bf 97.3\% & \,\  9.3\% \\ 
& \UAIpref & UAI  adv. loss + attr              & \okI, \okA                  &     65.0 &     68.0\% &     89.4\% &     92.9\% & \,\  9.5\% \\ 
& \Yref    & Adv. on $\vy$ only~\citep{Liu2018a}& \okI                        &     64.7 &     69.2\% &     83.6\% &     96.4\% & \,\  9.6\% \\ 
&          & \textbf{DualDis} (ours)            & \okI, \okA                  & \bf 68.0 &     71.1\% &     88.6\% & \bf 97.3\% & \bf 14.9\% \\ 

\midrule
\parbox[t]{3mm}{\multirow{8}{*}{\rotatebox[origin=c]{90}{\textbf{Yale-B}}}}

& \MTref   & Multi-task classif.                & \okI, \okA                  &     81.5 &     98.5\% &     97.2\% &     85.3\% &     45.1\% \\ 
& \HNref   & HybridNet-like~\citep{Robert2018}  & \okI                        &     65.3 &     97.6\% &     93.7\% &     23.3\% &     46.5\% \\ 
& \HNpref  & HybridNet-like + attr              & \okI, \okA                  &     80.5 & \bf 99.0\% &     96.9\% &     80.0\% &     46.1\% \\ 
& \MTANref & MTAN~\citep{Liu2018}               & \okI, \okA, \okAt           &      --  &     98.4\% &        --  &        --  &     70.3\% \\ 
& \UAIref  & UAI  adv. loss~\citep{Jaiswal2018} & \okI                        &     60.0 &     98.6\% &     65.5\% &     28.1\% &     48.0\% \\ 
& \UAIpref & UAI  adv. loss + attr              & \okI, \okA                  &     65.1 &     96.1\% &     95.8\% &     44.4\% &     24.1\% \\ 
& \Yref    & Adv. on $\vy$ only~\citep{Liu2018a}& \okI                        &     79.8 &     98.3\% &     84.1\% &     92.5\% &     44.4\% \\ 
&          & \textbf{DualDis} (ours)            & \okI, \okA                  & \bf 92.0 &     98.6\% & \bf 97.3\% & \bf 98.8\% & \bf 73.4\% \\ 

\midrule
\parbox[t]{3mm}{\multirow{8}{*}{\rotatebox[origin=c]{90}{\textbf{NORB}}}}

& \MTref   & Multi-task classif.                & \okI, \okA                  &     53.7 &     93.0\% &     84.2\% &     13.5\% &     24.0\% \\ 
& \HNref   & HybridNet-like~\citep{Robert2018}  & \okI                        &     51.1 &     93.3\% &     76.8\% &     12.2\% &     22.1\% \\ 
& \HNpref  & HybridNet-like + attr              & \okI, \okA                  &     52.5 &     92.9\% &     84.1\% &     10.7\% &     22.2\% \\ 
& \MTANref & MTAN~\citep{Liu2018}               & \okI, \okA, \okAt           &      --  &     92.2\% &        --  &        --  & \bf 30.5\% \\ 
& \UAIref  & UAI  adv. loss~\citep{Jaiswal2018} & \okI                        &     51.8 &     92.8\% &     76.0\% &     13.7\% &     24.7\% \\ 
& \UAIpref & UAI  adv. loss + attr              & \okI, \okA                  &     52.5 &     93.2\% &     82.8\% & \,\  8.0\% &     26.0\% \\ 
& \Yref    & Adv. on $\vy$ only~\citep{Liu2018a}& \okI                        &     67.3 &     92.2\% &     76.9\% &     78.9\% &     21.1\% \\ 
&          & \textbf{DualDis} (ours)            & \okI, \okA                  & \bf 72.3 & \bf 93.5\% & \bf 84.5\% & \bf 80.7\% & \bf 30.5\% \\ 

        \bottomrule
    \end{tabular}
\end{table}

\section{Experiments}
\label{sec:exp}

We validate our model on three datasets preprocessed to fit our training protocol: CelebA~\citep{celeba}, Yale-B~\citep{yale} and NORB~\citep{norb}. CelebA is a face dataset, we use 60k images with 2000 identities and 40 attributes (hair style, makeup, expression, \textit{etc.}). Yale-B is also a face dataset of 2.4k images with 38 identities and 14 attributes (light source position). NORB is a dataset of 48k images of 3D object renderings with 5 categories (10 objects per category) and 8 attributes (camera position and lighting).
We use regular convolutional architectures with batch normalization. Depending on the dataset, encoders range from 6 to 8 layers, decoders between 7 and 13 layers and latent spaces $\vhy$ and $\vhz$ are 80 to 196 units. Our model is not really sensitive to the hyperparameters and the adversarial training does not present instabilities like can be encounters with GANs. All the details about data preprocessing, train/validation/test splitting, model architectures, training details and hyperparameters are provided in the supplementary.

\subsection{Disentangling evaluation}
\label{sec:exp_dis}

\begin{figure}[tb]
    \centering
    \includegraphics[width=\linewidth]{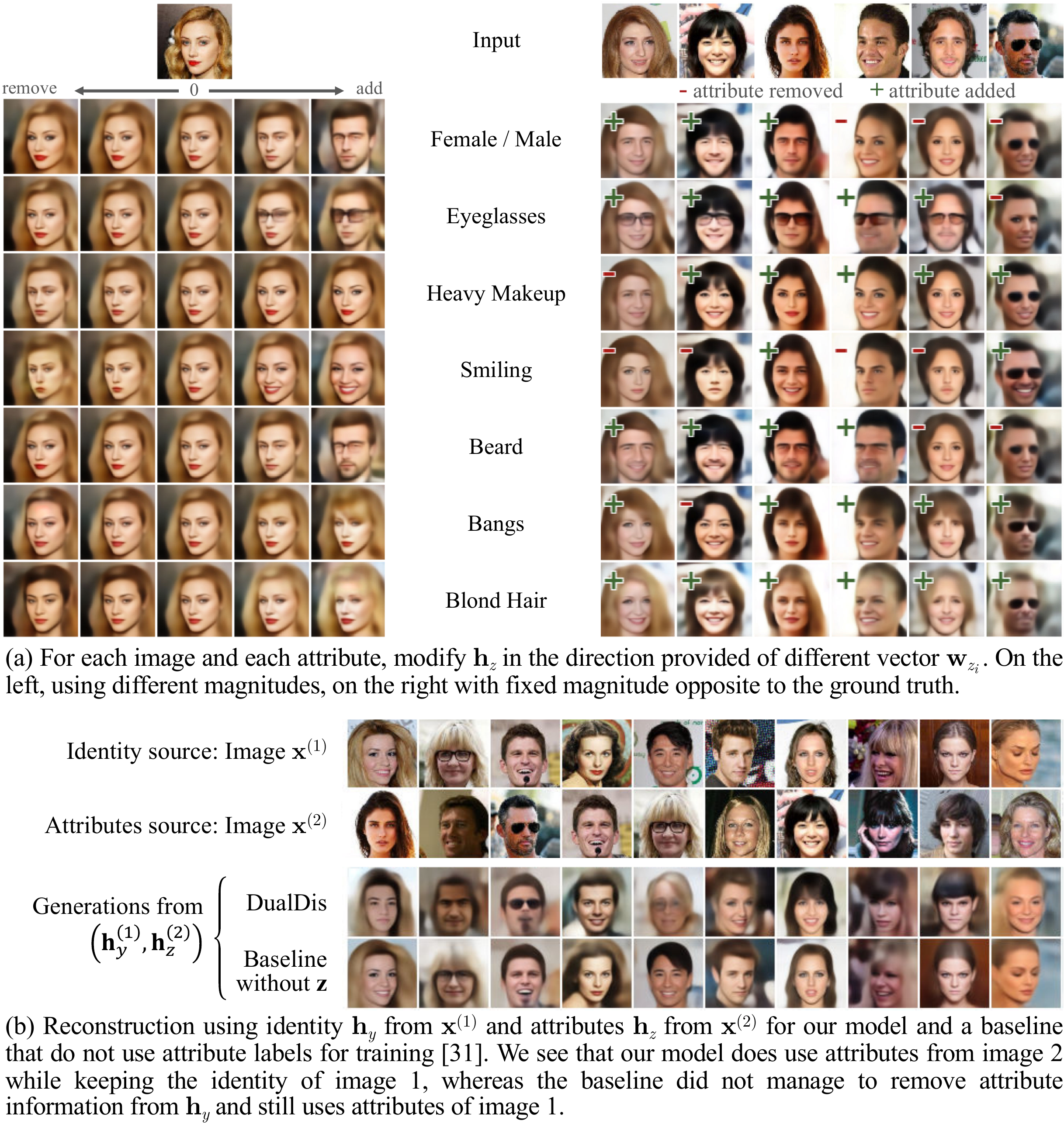}
    \caption{Visualizations of image editing on CelebA.}
    \label{fig:celeba}
\end{figure}

First, we demonstrate the ability of DualDis to successfully disentangle and linearize the factors of variation of the two information domains (class/identity and attributes) in the latent spaces.
We compare our model to baselines described in Sec.~\ref{sec:discussion} that are reproduced similarly to an ablation study by deactivating components of our model or by making small changes, \textit{cf.} Fig.~\ref{fig:archi}. We can therefore compare our model to fair reimplementations of the state of the art, and also evaluate variants of the baselines with labels for both information domains like DualDis. All models have the same architecture (in $E$, $E_y$, \textit{etc.}) and hyperparameters whenever those are common between models. Only UAI~\citep{Jaiswal2018} has a slightly different architecture since it requires shallow encoders $E_y$ and $E_z$.

To evaluate the representation and disentangling of the information domains, we measure the \textit{accuracy} of the linear classifiers $\vW_y$ and $\vW_z$ which indicates if the information is correctly extracted and linearized; and the \textit{error rate} (100 - accuracy) of the adversarial classifiers $C_y$ and $C_z$, to measure how much of the ``undesired'' information has been removed through disentangling\footnote{For models that do not include classifiers $C$, we add and train them to obtain disentangling metrics, without impacting the behavior of the rest of the model, \textit{cf.} details in the supplementary.}. To summarize the overall quality of a model, we propose an \textit{aggregated metric} that averages those 4 values.

Results are presented in Table~\ref{tab:sota} with baselines labeled (A) to (E).
DualDis provides the best performances on the three datasets with a gain of 3 to 10\,pts in the aggregated metric compared to the best baseline. We obtain strong disentangling results while having similar or better classification accuracies.
Looking at the baselines, first, the multi-task classifier \MTref~\citep{kokkinos2017} provides good accuracies but completely fails at disentangling the information and does not allow generation, making classification easier.
Adding reconstruction \HNrefs~\citep{Robert2018} regularizes the model and slightly helps the disentangling metrics by grounding latent features to specific visual patterns in the reconstruction. In comparison, DualDis provides an explicit disentangling mechanism to greatly improve disentangling metrics.
MTAN~\MTANref produces good results but uses a single latent space $\vhy$ and a specific method always requiring labels $\vz$ as inputs of the decoder to be used, even in test, making it hard to use. This constraint causes lower quality reconstructions and also limits the comparison to DualDis.
The orthogonalization mechanism on latent representations proposed by UAI~\UAIrefs produces weak disentangling results. Finally, the asymmetrical disentangling of \Yref~\citep{Liu2018a}, only applied to $\vy$, lacks the ability to remove the attribute information in the identity space.
In comparison, we combine reconstruction for generation and features regularization; classification and linearization to improve the semantic quality of the features; and leverage a symmetrical label-guided adversarial disentangling to effectively separate the two domains in both latent spaces.

\parag{Semi-supervised learning (SSL).} In addition, using SSL, we validate our ability to provide disentangling and linearization with few attribute labels, relaxing the constraint of having labels for both domains. On CelebA with 2\% of labels we obtain an aggregated score of 65.5, and with 4\% we obtain 66.8, better than the baselines in Table~\ref{tab:sota}. We also confirm the possibility of semantic image manipulation similarly to the full model, shown in the supplementary along with more results.

\subsection{Applications of DualDis}

We now evaluate two possible applications of this architecture: image editing on CelebA and image generation for data augmentation on Yale-B.

\parag{Image manipulation.} Using the disentangling and linearization abilities of DualDis, we perform semantic image manipulation with linear changes of $\vhz$ along attribute directions $\vw_{z_i}$, \textit{cf.} Sec.~\ref{sec:model_factors}. On Fig.~\ref{fig:celeba}a (left) we do this for different attributes and move in positive and negative directions with different magnitudes. We can finely control the importance of an attribute, for example adding a small or big smile, different shades of blonde, \textit{etc.} We then fix a reasonable magnitude threshold for when an attribute changes enough and apply the same technique with this threshold to flip the the ground truth labels of different images, on the right of Fig.~\ref{fig:celeba}a. We can switch the gender, add and remove eyeglasses, bangs, \textit{etc.} This shows that we effectively modeled and linearized attribute information in $\vhz$. Visualizations for Yale-B and NORB are available in the supplementary.

We can also mix representations of different images. In particular, we propose to use $\vhy$ from a first image and $\vhz$ from a second, and reconstruct an image from those. We do this for DualDis and the baseline not using attributes \Yref~\citep{Hadad2018,Liu2018a}. This experiment is a qualitative verification that our model indeed separated the information of both domains. Results are presented on Fig.~\ref{fig:celeba}b and show that our model is able to mix the identity of the first image and the attributes of the second, while the baseline was not able to correctly separate the two types of information and uses attributes from the first image. For example, for the first pair, we keep the original identity but change the smile and hair color; for the second we change the gender and glasses; for the third we add glasses, \textit{etc.} This confirms that our symmetrical disentangling on both domains is able to separate the two types of information that otherwise remain entangled for a complex dataset like CelebA.

\begin{wraptable}{r}{0.53\textwidth}
\setlength{\tabcolsep}{4pt}
\begin{tabular}{@{}cccccc@{}}
        \toprule
        Initial     & \multicolumn{5}{c}{Nb. generated images per class}  \\
        \cmidrule{2-6}
        train size  & 0 & 10 & 20 & 30 & 60 \\
        \midrule
        480 & 78.9\% & 79.3\% & 80.1\% & 81.6\% & \textbf{82.8\%} \\
        360 & 69.1\% & 70.5\% & 72.6\% & 73.1\% & \textbf{75.6\%} \\
        240 & 48.9\% & 51.8\% & 55.5\% & 56.8\% & \textbf{58.6\%} \\
        \bottomrule
    \end{tabular}
    \caption{Accuracy of identity prediction on Yale-B using generated images as data augmentation.
\label{tab:DA}}
\end{wraptable}

\parag{Image generation for data augmentation.} Finally, we demonstrate on Yale-B our ability to produce new samples for semantic guided data augmentation. Since Yale-B contains different lighting for each identity, when using a restricted number $N_\textrm{init}$ of images in train, each class contains only a small portion of the possible lighting variations. We first train a DualDis model with $N_\textrm{init}$ training images. Based on the linearization and editing properties of our model, we generate variations in attributes (similarly to Fig.~\ref{fig:celeba}a) from train images to obtain new images with the same identity but a new and known attribute label $\vz'$. For each identity, we generate $N_\textrm{gen}$ new images,
obtaining images with attributes missing in the original train set as well as increasing the number of instances of existing attributes for robustness. This provides a more representative dataset of the variations in attributes for each identity.

We apply this procedure for different values of $N_\textrm{init}$ and $N_\textrm{gen}$. We then learn a classifier
on a train set contains original and generated images and evaluate its accuracy on the test set. The results are reported in Table~\ref{tab:DA}. Adding generated images to the train set with new attribute variations for each identity provides a gain in our 3 setups. When the initial train set is small, the gain is larger, with a gain of almost 10\,pts between the baseline without data augmentation and a data augmentation of 60 images per class. Adding more than 60 images per class does not yield a significant improvement.

\section{Conclusion}

In this paper, we present DualDis, a disentangling model designed to effectively separate two information domains in two latent subspaces. Using labels for both domains, we obtain improved classification and disentangling results while linearizing factors of variation; and validate the possibility of reducing annotations requirements using semi-supervised learning with maintaining good disentangling properties.
Using our structured latent space, we are able to perform image editing by changing attributes on CelebA. We also carry out semantic data augmentation on Yale-B and obtain important identity classification improvements.
In future work, it would be interesting to bring this model closer to generative models, improving image quality and allowing generation from noise while maintaining a structured latent space. %

\FloatBarrier

\bibliography{biblio}

\newpage
\setcounter{page}{1}

\makeatletter
\vbox{%
    \hsize\textwidth
    \linewidth\hsize
    \vskip 0.1in
    \@toptitlebar
    \centering
    {\LARGE\bf \@title\newline \Large \textit{Supplementary material} \par}
    \@bottomtitlebar
    \vskip 0.3in \@minus 0.1in
  }
\makeatother
\FloatBarrier

\appendix
\raggedbottom

\section*{Table of contents}
\startcontents[sections]
\printcontents[sections]{l}{1}{\setcounter{tocdepth}{2}}

\section{Data pre-processing}

\subsection{CelebA}

The official CelebA dataset\footnote{\url{http://mmlab.ie.cuhk.edu.hk/projects/CelebA.html}} contains $\sim$200K images for 10,177 identities. As is common, we used the cropped and aligned version. However, the number of images per identity varies and some have very few images. Since our purpose is to work on datasets with two classification tasks, we chose to reduce the number of identities to 2,000, keeping those with the highest number of images. Because of this, we obtain a dataset with $\sim$60K images.

The identities are our label $\vy$ and the attributes provided with the dataset are were not preprocessed and are used as $\vz$.

\subsection{Yale}

The Extended Yale-B dataset\footnote{\url{http://vision.ucsd.edu/~leekc/ExtYaleDatabase/ExtYaleB.html}} is available for download in two variants: the ``full'' version contains 16128 images of 28 human subjects under 9 poses and 64 illumination conditions, each image has a large part of background; and the ``cropped'' version contains $\sim$2400 images with 38 subjects and 64 illumination conditions with no background. We chose to work with the cropped version.

The 38 identities constitute our identity label $\vy$. The lighting source information is provided as 2 real values indicating the angles (elevation and azimuth) of the light source. We propose to convert this information in 14 ``clusters'', we show the id of each cluster between 0 and 13 in this table:
\begin{center}
    \includegraphics[width=0.35\textwidth]{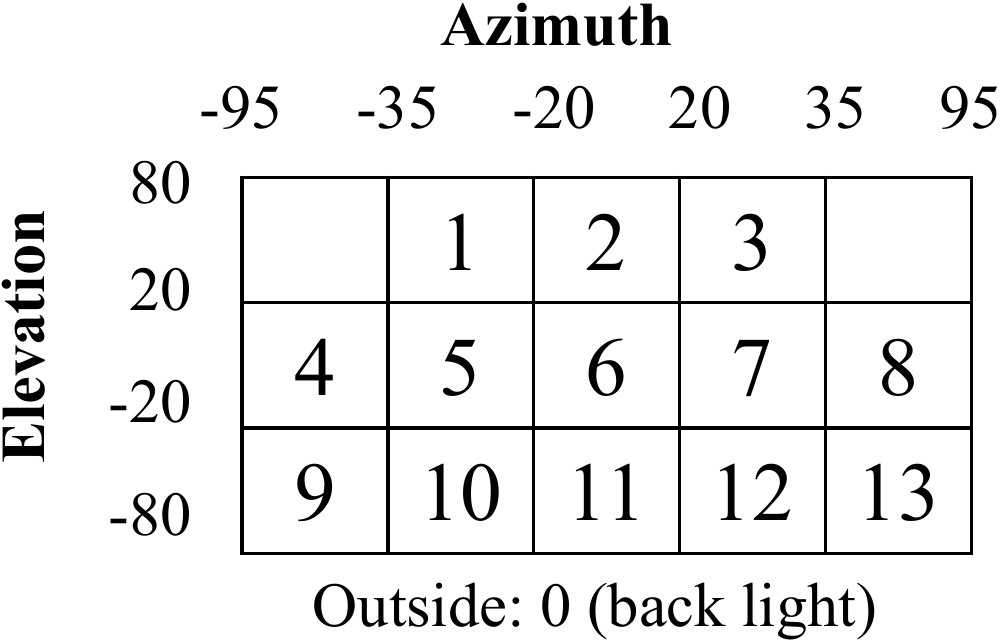}
\end{center}
Each image is attributes to one of the clusters, and the label $\vz$ is a one-hot vector indicating the lighting source.

\subsection{NORB}

The NORB dataset\footnote{\url{https://cs.nyu.edu/~ylclab/data/norb-v1.0/}} ``contains images of 50 toys belonging to 5 generic categories: four-legged animals, human figures, airplanes, trucks, and cars. The objects were imaged by two cameras under 6 lighting conditions, 9 elevations (30 to 70 degrees every 5 degrees), and 18 azimuths (0 to 340 every 20 degrees).'' We use the base dataset without jitter. The 5 categories are used as our classes $\vy$, and a process similar to Yale-B is used to create the labels $\vz$, but with soft assignment:
\begin{itemize}
	\item The 6 lighting classes are converted into a single unit with values between 0 (dark) and 1 (very light) with mapping as follows: [0.6, 0.3, 0, 0.7, 0.4, 1]
	\item The elevation is represented by 3 clusters $e_i$ of centers [35, 50, 65] with an assignment to each defined as $\vz_i = 1 - \min(1, |elevation - e_i| / 15)$
	\item The azimuth is represented by 4 clusters $a_j$ of center [0, 90, 180, 270] with an assignment to each defined as $\vz_j = 1 - \min(1, |azimuth - a_j| / 9)$
\end{itemize}
This gives us a complete vector $\vz$ of size 8.

\subsection{Dataset splitting}

CelebA and Yale have no provided test set for classification. Therefore, we create a test set by using $x\%$ of the images of each class, regardless of the attributes. For CelebA, we use 20\% of the images and for Yale-B 50\% of the images. For NORB, the test set is provided, it consists of 25 of the 50 3D objects, and represents 50\% of the images.

The validation set represents 20\% of the training set, defined with the same process, and is used for the few hyper-parameters tests we do.

\section{Architectures \& Hyperparameters}

\subsection{Complete architecture overview}

\begin{figure}[H]
    \centering
    \includegraphics[width=\textwidth]{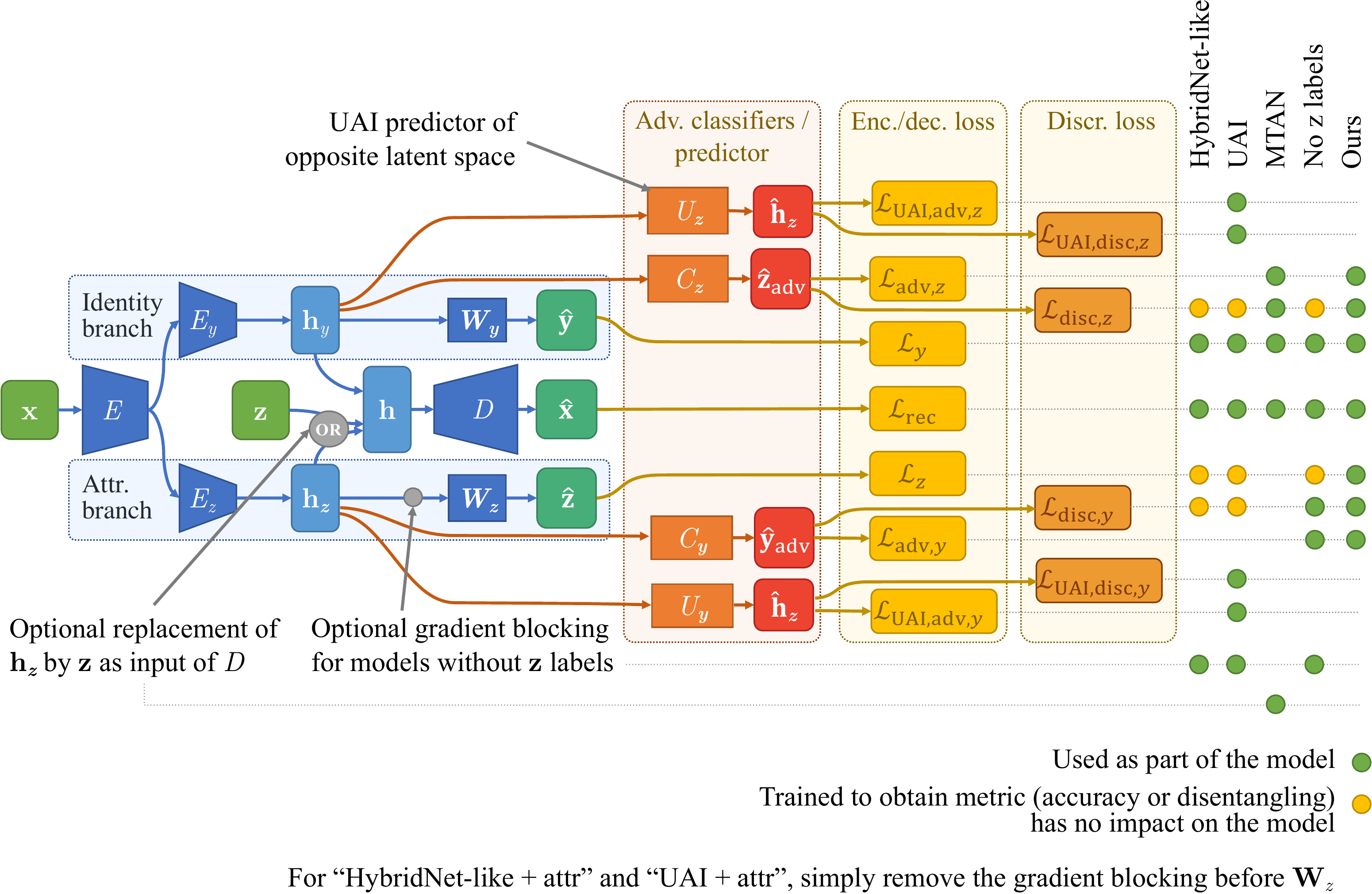}
    \caption{Complete architecture with all possible options and variants. We also show which loss terms are activated to reproduce the different models that we report.}
    \label{fig:archis}
\end{figure}

To obtain the different models that we report, we use start from a complete architecture with all possible options, and then choose which parts of the model we activate. This complete architecture is represented on Fig.~\ref{fig:archis}.

In particular, we can see that we have:
\begin{itemize}
    \item $U_y$ and $U_z$ which are the predictors that replace $C_y$ and $C_z$ for UAI model, and that predict $\vhy$ and $\vhz$. Their corresponding loss term is a MSE for both $\calL_\textrm{UAI,adv}$ and $\calL_\textrm{UAI,disc}$ that is maximized for $\calL_\textrm{UAI,adv}$ and minimized for $\calL_\textrm{UAI,disc}$.
    \item The possibility to replace $\vhz$ by $\vz$ as the second input of the decoder $D$, which allows to produce MTAN.
    \item The possibility to block gradients between $\vhz$ and $\vW_z$, which means that when activated we can train $\vW_z$ to measure the quality of the representation $\vhz$ regarding the attributes, while not backpropagating this signal to $E_z$, therefore reproducing models that do not use $\vz$ labels while keeping the metric.
    \item It is possible to train adversarial classifiers $C_y$ and $C_z$ with $\calL_\textrm{disc}$ even when not proposed by the models, which allows to measure the quality of the disentangling and will have no impact of the actual model as long as $\calL_\textrm{adv}$ are disabled since $\calL_\textrm{disc}$ is only backpropagated in $C$.
\end{itemize}

\subsection{Architecture details}

In Table~\ref{tab:archis}, we provide the exact details about the architecture of the components in Fig.~\ref{fig:archis} depending on the experiments and the dataset.

\newcommand{\nlspace}{\rule{0pt}{9pt}}
\newcommand{\li}{$\ell$}
\begin{table}[htbp]
    \centering
    \begin{tabular}{r|p{0.41\textwidth}|p{0.41\textwidth}}
        \toprule
         & Architecture for UAI & Architecture for all other models \\
        \midrule
        \multicolumn{3}{c}{\textbf{CelebA} (image size 256$\times$256)}\\
        \midrule
        
       $E$
       & 32p0s2, 32p0s1, 64p0, maxpool2k3, 80k1, maxpool2k3, 96p0, maxpool2k3, 128p0, 160p0s2, 196p0
       & 32p0s2, 32p0s1, 64p0, maxpool2k3, 80k1, maxpool2k3, 96p0, maxpool2k3
       \\ \nlspace
       $E_y/E_z$
       & 196p0
       & 96p0, 128p0s2, 196p0, 196p0
       \\ \nlspace
       $D$
       & \multicolumn{2}{p{0.82\textwidth}}{dec392p0s1, 392, upsample, 392, upsample, 256, upsample, 196, upsample, 128, 128, upsample, 96, 96, upsample, 64, 64, 32, 3k1}
       \\ \nlspace
       $\vW_y$ & \li2000 & \li2000 \\ \nlspace
       $\vW_z$ & \li40 & \li40 \\ \nlspace
       $C_y$ & \li256, \li2000 & \li256, \li256, \li2000 \\  \nlspace
       $C_z$ & \li256, \li40 & \li256, \li256, \li40 \\ \nlspace
       $U_y$ & \li196 & N/A \\\nlspace
       $U_z$ & \li196 & N/A \\

        \midrule
        \multicolumn{3}{c}{\textbf{Yale-B} (image size 64$\times$64)}\\
        \midrule
        
       $E$
       & 32k4s2,40k4s2,48k4s2,76k4s2,100k3p0
       & 32k4s2, 40k4s2, 48k4s2
       \\ \nlspace
       $E_y/E_z$
       & 80k2p0
       & 64k4s2, 72k3p0, 80k2p0
       \\ \nlspace
       $D$
       & \multicolumn{2}{p{0.82\textwidth}}{160k2p1, dec80, dec64, dec48, dec32, dec32, 32, 3none}
       \\ \nlspace
       $\vW_y$ & \li38 & \li38 \\ \nlspace
       $\vW_z$ & \li14 & \li14 \\ \nlspace
       $C_y$ & \li80, \li80, \li38 & \li80, \li80, \li38 \\  \nlspace
       $C_z$ & \li80, \li80, \li14 & \li80, \li80, \li14 \\ \nlspace
       $U_y$ & \li80 & N/A \\\nlspace
       $U_z$ & \li80 & N/A \\

        \midrule
        \multicolumn{3}{c}{\textbf{NORB} (image size 64$\times$64)}\\
        \midrule
        
       $E$
       & 64k4s2,64k4s2,96k4s2,164k4s2,192k4s2
       & 64k4s2, 64k4s2, 96k4s2
       \\ \nlspace
       $E_y/E_z$
       & 128k2p0
       & 96k4s2, 128k3p0, 128k2p0
       \\ \nlspace
       $D$
       & \multicolumn{2}{p{0.82\textwidth}}{256k2p1, dec192, 128, dec128, 128, dec96, 96, dec64, 64, dec64, 64, 32, 1}
       \\ \nlspace
       $\vW_y$ & \li5 & \li5 \\ \nlspace
       $\vW_z$ & \li8 & \li8 \\ \nlspace
       $C_y$ & \li128, \li128, \li5 & \li128, \li128, \li5 \\  \nlspace
       $C_z$ & \li128, \li128, \li8 & \li128, \li128, \li8 \\ \nlspace
       $U_y$ & \li128 & N/A \\\nlspace
       $U_z$ & \li128 & N/A \\
        
        \bottomrule
    \end{tabular}
    \caption{Architectures used for our experiments.\newline\newline
    In the \textbf{encoder}, every layer is followed by batch normalization and ReLU.\newline
    In the \textbf{decoder}, every layer is followed by a batch normalization and LeakyReLU(0.2), except last layer which has no activation or BN.\newline
    In the \textbf{classifiers}, every intermediate layer is followed by a ReLU.\newline
    \newline
    \underline{\textbf{Layers description syntax:}}\newline
    \textbf{Conv:} \texttt{128[k5][p0][s2]} is a convolutional layer with 128 output channels, a kernel of 5 (default is 3 if not written), padding of 0 (default is to keep same output size), stride 2 (default is 1)\newline
    \textbf{Deconv:} \texttt{dec128[k4][p0][s1]} is a transpose convolutional layer with 128 output channels, a kernel of 4 (default), padding of 0 (default is 1), stride 1 (default is 2)
    \textbf{Linear:} \li\texttt{128} is a linear layer with 128 output neurons\newline
    \textbf{Upsample:} \texttt{upsample} means an upsampling of a factor 2 using the nearest value  \newline
    \textbf{MaxPool:} \texttt{maxpool2k3} is a max pooling of stride 2 and kernel 3
    }
    \label{tab:archis}
\end{table}

\subsection{Training and hyperparameters values}

As a reminder, we have two losses composed of different loss terms, each weighted by a parameter $\lambda$ that controls its importance. Here is the global loss:
\begin{align}
\calL &= \lambda_{rec} \calL_\textrm{rec} + \lambda_y \calL_{y} + \lambda_z \calL_{z} + \lambda_{adv,y} \calL_{\textrm{adv},y} + \lambda_{adv,z} \calL_{\textrm{adv},z} + \lambda_{o} \calL_\textrm{orth}\,.\\
\calL_\mathrm{disc} &= \lambda_{disc,y} \calL_{\textrm{disc},y} + \lambda_{disc,z} \calL_{\textrm{disc},z}\,.
\end{align}

This loss is optimized using Adam with the recommended hyperparamters: learning rate of $0.001$, $\beta_1 = 0.9$, $\beta_2 = 0.999$. The number of epochs and batch sizes depends on the datasets and are given below.

For all the experiments, we used $\lambda = 1$ for all the classification losses:
$\lambda_y = \lambda_z = \lambda_{disc,z} = \lambda_{disc,z} = \lambda_{UAI,disc,z} = \lambda_{UAI,disc,z} = 1$, and we set $\lambda_{o} = 1\times 10^{-6}$

Values of hyperparameters that depends on the dataset are provided in Table~\ref{tab:HP}.

\begin{table}[H]
    \centering
    \begin{tabular}{rcccccccccccc}
        \toprule
        & $\lambda_{rec}$ & $\lambda_{adv,y}$ & $\lambda_{adv,z}$ & $\lambda_{UAI,adv,y}$ & $\lambda_{UAI,adv,z}$ & BS & Epochs \\
        \midrule
        CelebA       & 0.3 & 0.1  & 0.1  & 0.3 & 0.3 & 32 & 330 \\
        Yale          & 1   & 0.08 & 0.08 & 0.3 & 0.3 & 64 & 400 \\
        NORB         & 10  & 0.25 & 0.25 & 0.3 & 0.3 & 128 & 250 \\
        \bottomrule
    \end{tabular}
    \caption{Hyperparameters for the various experiements. BS = Batch Size.}
    \label{tab:HP}
\end{table}

\section{Experiments details}

\subsection{Image editing}

For image editing, we use the model trained for the ablation study of the datasets, and start by obtaining the representations $\vhy$ and $\vhz$ for some test images. For attribute modification, we move $\vhz$ in the direction $i$ of each vector $\vw_{z_i}$ to obtain $\vhz' = \vhz \pm \varepsilon \vw_{z_i}$ of the model and produce images using the decoder: $\vxh = D(\vhy, \vhz')$. The amplitude of $\varepsilon$ for the visualization was fixed after a quick visual check of what values looked.
For identity / attributes mixing between images, we simply use $\vhy$ and $\vhz$ from different images and feed them to the decoder.

\subsection{Semi-supervised learning}

For semi-supervised learning, we use batches with a pre-defined number of supervised images in each batch. We iterate over the set of labeled and unlabeled images independently and consider an epoch as the loop over the unlabeled image set, during which we usually see images of the supervised set more than once depending on the size of the sets. This is a common setting, e.g.~\cite{Sajjadi2016,tarvainen2017mean,Robert2018}. The hyperparameters that differ from the model in the ablation study are provided in Table~\ref{tab:HP_SSL}.

\begin{table}[H]
    \centering
    \begin{tabular}{rcccccccccccc}
        \toprule
        & $\lambda_{rec}$ &  $\lambda_{z}$ & $\lambda_{adv,y}$ & $\lambda_{adv,z}$ & Sup BS \\
        \midrule
        4000     & 0.3 & 0.4 & 0.2  & 0.1 & 10 \\
        2000     & 0.5 & 0.4 & 0.2  & 0.1 & 10 \\
        1000     & 0.5 & 0.4 & 0.2  & 0.1 & 8 \\
        400      & 0.5 & 0.4 & 0.2  & 0.1 & 8 \\

        \bottomrule
    \end{tabular}
    \caption{Hyperparameters for the SSL experiements. Sup BS = Number of labeled images in each batch.}
    \label{tab:HP_SSL}
\end{table}

\subsection{Data augmentation on Yale}

For the data augmentation experiments, we start by training a new model with different sizes of train datasets. Once trained, we produce 150 new images for each identity using the image editing technique we described in order to produce new images of the different attribute categories. For this, we iterate over the train images of each identity (after excluding train images with attributes that correspond to very bad lighting, \textit{i.e.} attributes 0, 4, 8, 9, 13) and then randomly choose an attribute for which we do not already have enough images for this identity. This is done so that after data augmentation, each identity has images that follows this distribution $\mathcal D$ over attributes 0 to 13:
\begin{equation}
    \mathcal D = [1, 3, 3, 2, 5, 3, 10, 3, 5, 2, 2, 2, 2, 2] / 45
\end{equation}

Then, a classifier with the architecture $\vW_y \circ E_y \circ E$ is trained with Adam for 400 epochs on the original train set used for to train the generator + the generated images. It is then evaluated on all the remaining images of the original dataset.

\section{Additional results}

\subsection{Image editing on CelebA}

\begin{figure}[H]
    \centering
    \includegraphics[width=\linewidth]{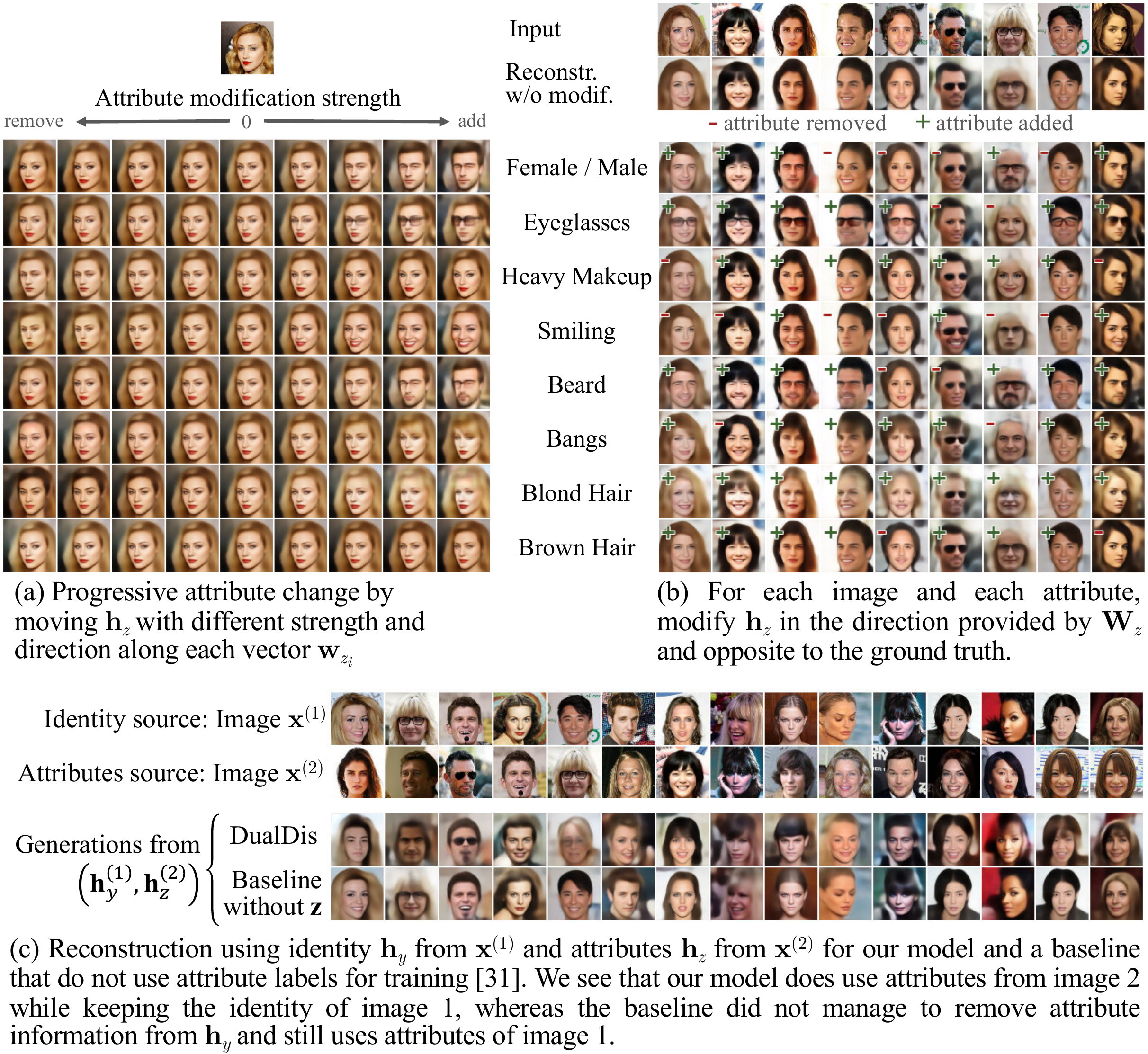}
    \caption{More detailed visualizations of image editing on CelebA}
    \label{fig:celebafull}
\end{figure}

\subsection{Image editing on Yale and NORB}

\begin{figure}[H]
    \centering
    \includegraphics[width=0.75\linewidth]{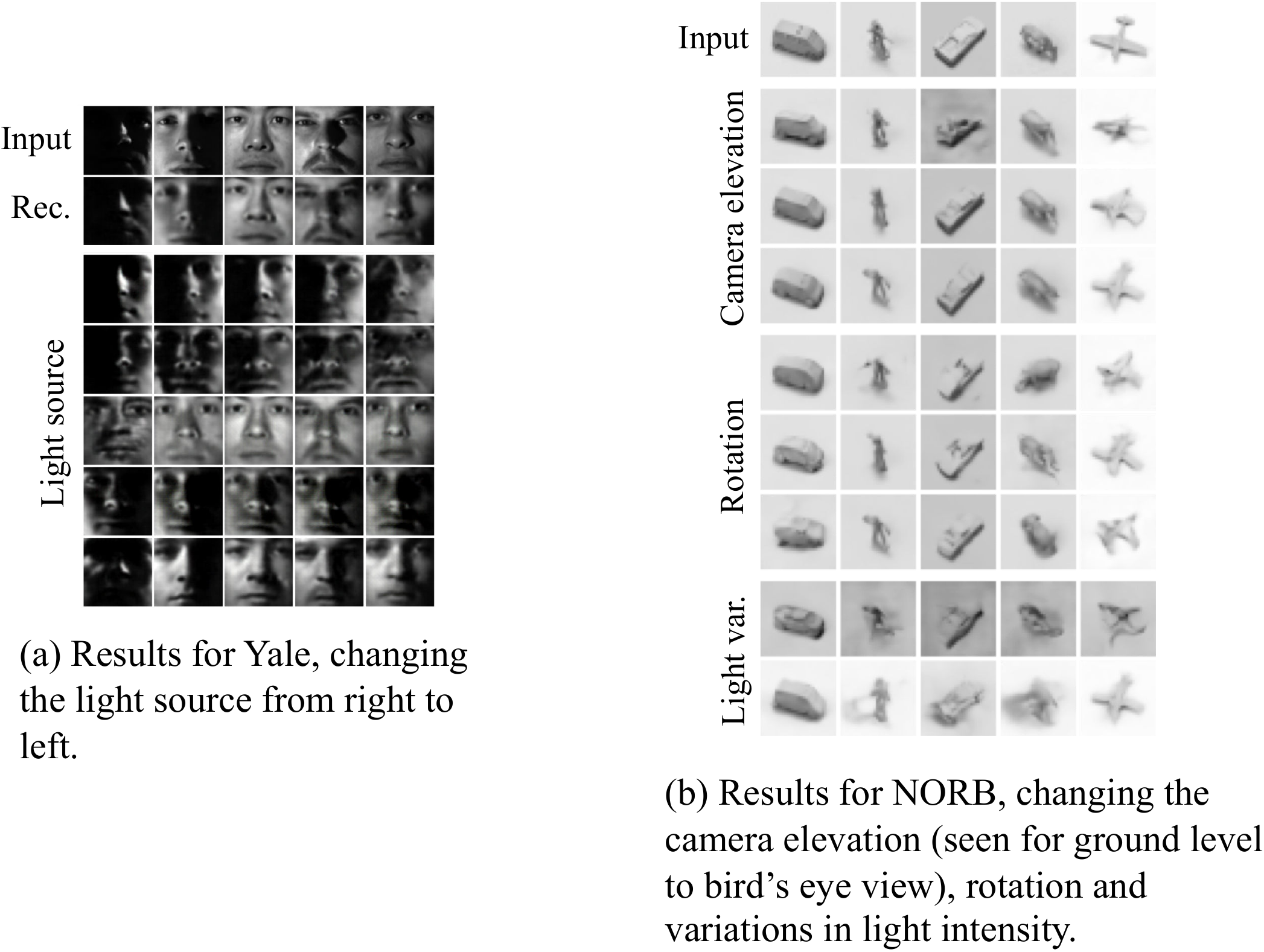}
    \caption{Visualizations of image editing on Yale-B and NORB}
    \label{fig:yale_norb}
\end{figure}

\subsection{Semi-supervised learning}

\begin{figure}[H]
    \centering
    \includegraphics[width=0.7\textwidth]{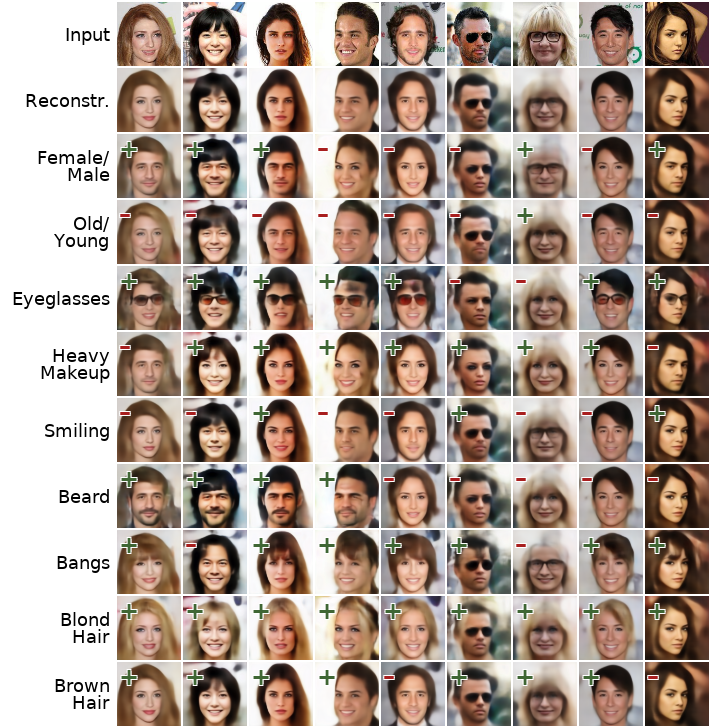}
    \caption{Image editing using our model trained with semi-supervised learning with 1000 attributes labels. Equivalent of Fig.~\ref{fig:celebafull}b.}
    \label{fig:SSL_edit}
\end{figure}

For the semi-supervised experiment, we tried both using a simple semi-supervision by ignoring unlabeled samples when labels were needed, and the more advanced Mean Teacher~\citep{tarvainen2017mean} strategy but found small differences, which might be due to a sub-optimal tuning of the sensitive hyper-parameters of Mean Teacher. We therefore report results for the simple solution in Table~\ref{tab:SSL}. We can see that even 1000 or 2000 labels produce reasonable results that are fairly close to the model trained with full supervision. In addition, we confirmed via a qualitative analysis of the model with 1000 labels that it can produce image edition with results similar to the fully supervised models. This is shown in Fig.~\ref{fig:SSL_edit} and was generated in the same way as for the full model in Fig.~\ref{fig:celeba}.

\begin{table}[H]
\caption{Results of disentangling on CelebA using semi-supervised learning on attribute labels. 48k labels is the fully supervised baseline.}
\label{tab:SSL}
\setlength{\tabcolsep}{4pt}
    \begin{tabular}{@{}r@{\hspace{4pt}}c@{\hspace{4pt}}cc@{\hspace{4pt}}cc@{}}
        \toprule
        Nb. attr.        & Aggr.                             &  \multicolumn{2}{c}{Accuracy} &  \multicolumn{2}{c}{Disentangling} \\
        labels        & metric                & ${\scriptstyle \vh_{\scriptstyle y} \rightarrow} \vy$
                                    & ${\scriptstyle \vh_{\scriptstyle z} \rightarrow} \vz$
                                    & ${\scriptstyle \vh_{\scriptstyle z} \rightarrow} \vy_\textrm{adv}$
                                    & ${\scriptstyle \vh_{\scriptstyle y} \rightarrow} \vz_\textrm{adv}$ \\
\toprule
          400   &     63.9 &     65.2\% &     81.2\% & \bf 97.7\% &     11.6\% \\ 
         1000   &     65.5 &     68.4\% &     84.3\% &     97.4\% &     11.9\% \\ 
         2000   &     66.8 &     71.0\% &     85.0\% & \bf 98.4\% &     12.7\% \\ 
         4000   &     67.6 & \bf 72.6\% &     85.8\% & \bf 98.3\% &     13.8\% \\ 
         48000  & \bf 68.0 &     71.1\% & \bf 88.6\% &     97.3\% & \bf 14.9\% \\ 
        \bottomrule
    \end{tabular}
\end{table}

\end{document}